\DocumentMetadata{}
\documentclass[sigconf]{acmart}
\usepackage{hyperref}
\usepackage{hyperxmp}
\usepackage{tikz}
\usepackage{pifont}
\usepackage{wasysym}
\usepackage{mathtools}
\usepackage{lipsum}
\usepackage{graphicx}
\usepackage{caption, subcaption}
\usepackage{url}
\usepackage{amsmath, amsfonts, amsthm, bm}
\usepackage{soul}
\usepackage{tabularx, multirow}
\usepackage{dsfont}
\usepackage[ruled,vlined]{algorithm2e}
\usepackage{booktabs}
\usepackage{xcolor, colortbl}
\usepackage{xspace}
\usepackage{resizegather}
\usepackage{makecell}
\usepackage{CJKutf8}
\usepackage{enumitem}

\newcolumntype{P}{>{\raggedright\arraybackslash}m{0.95\linewidth}}
\newcolumntype{Q}{>{\centering\arraybackslash}m{0.11\linewidth}}
\newcolumntype{R}{>{\raggedright\arraybackslash}m{0.12\linewidth}}
\newcolumntype{S}{>{\centering\arraybackslash}m{0.06\linewidth}}
\newcolumntype{U}{>{\centering\arraybackslash}m{0.12\linewidth}}

\newcommand{\midsepremove}{\aboverulesep=1pt \belowrulesep=1pt}
\midsepremove
    
\AtBeginDocument{%
  }

\copyrightyear{2025} 
\acmYear{2025} 
\setcopyright{rightsretained} 
\setcctype{by} 
\acmConference[SIGIR '25]{Proceedings of the 48th International ACM SIGIR Conference on Research and Development in Information Retrieval}{July 13--18, 2025}{Padua, Italy}
\acmBooktitle{Proceedings of the 48th International ACM SIGIR Conference on Research and Development in Information Retrieval (SIGIR '25), July 13--18, 2025, Padua, Italy}
\acmDOI{10.1145/3726302.3730055} 
\acmISBN{979-8-4007-1592-1/2025/07}

\begin{document}

\definecolor{DarkGreen}{RGB}{30,130,30}
\newcommand{\cmark}{\textcolor{DarkGreen}{\ding{51}}}
\newcommand{\xmark}{\textcolor{red}{\ding{55}}}%

\definecolor{DarkYellow}{RGB}{255,204,0}
\newcommand{\ytriangle}{\textcolor{DarkYellow}{\LARGE$\triangle$}}

\newcommand{\userset}{$\mathcal{U}$\xspace}           
\newcommand{\itemset}{$\mathcal{I}$\xspace}           
\newcommand{\interactions}{$\mathcal{R}$\xspace}           
\newcommand{\reviews}{$\mathcal{V}$\xspace}           
\newcommand{\predrating}{$\hat{r}_{u,i}$\xspace}
\newcommand{\actualrating}{${r}_{u,i}$\xspace} 
\newcommand{\useritempair}{$(u, i)$\xspace}           
\newcommand{\predfunc}{$f(u, i)$\xspace}        

\newcommand{\imdb}{IMDB\xspace}
\newcommand{\amazon}{Amazon-Book\xspace}

\newcommand{\mf}{MF\xspace}
\newcommand{\bprmf}{BPR-MF\xspace}
\newcommand{\mlp}{WDL\xspace}
\newcommand{\lightgcn}{LightGCN\xspace}
\newcommand{\rgcl}{RGCL\xspace}

\newcommand{\tallrec}{TALLRec\xspace}
\newcommand{\collm}{CoLLM\xspace}
\newcommand{\genrec}{GenRec\xspace}
\newcommand{\llamarec}{LlamaRec\xspace}
\newcommand{\llmrec}{LLMRec\xspace}
\newcommand{\pfive}{P5\xspace}
\newcommand{\pepler}{PEPLER\xspace}
\newcommand{\xrec}{XRec\xspace}
\newcommand{\trsr}{LLM-TRSR\xspace}
\newcommand{\allmrec}{A-LLMRec\xspace}
\newcommand{\llara}{LLaRA\xspace}
\newcommand{\slim}{SLIM\xspace}

\newcommand{\zsranker}{ZS-Ranker\xspace}
\newcommand{\gptthree}{GPT-3.5\xspace}
\newcommand{\gptfour}{GPT-4\xspace}
\newcommand{\proposed}{\textsc{Exp3rt}\xspace}
\newcommand{\proposednospace}{\textsc{Exp3rt}}

\newcommand{\userside}{user persona\xspace}
\newcommand{\itemside}{item synopsis\xspace}

\newcommand{\reduce}[1]{\textls[-50]{#1}}
\newcommand{\smallsection}[1]{{\vspace{0.05in} \noindent \bf {#1.\hspace{5pt}}}}




\title{Review-driven Personalized Preference Reasoning with Large Language Models for Recommendation}




\author{Jieyong Kim}
\authornote{Both authors contributed equally to this research.}
\affiliation{%
  \institution{Yonsei University}
  \city{Seoul}
  \country{Republic of Korea}}
\email{jieyong99@yonsei.ac.kr}

\author{Hyunseo Kim}
\authornotemark[1]
\affiliation{%
  \institution{Yonsei University}
  \city{Seoul}
  \country{Republic of Korea}}
\email{hyunseo00@yonsei.ac.kr}

\author{Hyunjin Cho}
\affiliation{%
  \institution{Yonsei University}
  \city{Seoul}
  \country{Republic of Korea}}
\email{cyberhyunjin@yonsei.ac.kr}

\author{SeongKu Kang}
\affiliation{%
  \institution{Korea University}
  \city{Seoul}
  \country{Republic of Korea}}
\email{seongkukang@korea.ac.kr}

\author{Buru Chang}
\affiliation{%
  \institution{Korea University}
  \city{Seoul}
  \country{Republic of Korea}}
\email{buru_chang@korea.ac.kr}

\author{Jinyoung Yeo}
\affiliation{%
  \institution{Yonsei University}
  \city{Seoul}
  \country{Republic of Korea}}
\email{jinyeo@yonsei.ac.kr}

\author{Dongha Lee}
\authornote{Corresponding author}
\affiliation{%
  \institution{Yonsei University}
  \city{Seoul}
  \country{Republic of Korea}}
\email{donalee@yonsei.ac.kr}

\renewcommand{\shortauthors}{Kim et al.}

\begin{abstract}
Recent advancements in Large Language Models (LLMs) have demonstrated exceptional performance across a wide range of tasks, generating significant interest in their application to recommendation systems.
However, existing methods have not fully harnessed the potential of LLMs, often constrained by limited input information or failing to fully utilize their advanced reasoning capabilities.
To address these limitations, we introduce \proposed, a novel LLM-based recommender designed to leverage rich preference information contained in user and item reviews.
\proposed is basically fine-tuned through distillation from a teacher LLM to perform three key steps in order:
(1) \textit{preference extraction},
(2) \textit{profile construction}, and
(3) \textit{textual reasoning for rating prediction}.
\proposed first extracts and encapsulates essential subjective preferences from raw reviews, next aggregates and summarizes them according to specific criteria to create user and item profiles.
It then generates detailed step-by-step reasoning followed by predicted rating, i.e., reasoning-enhanced rating prediction, by considering both subjective and objective information from user/item profiles and item descriptions.
This personalized preference reasoning from \proposed enhances rating prediction accuracy and also provides faithful and reasonable explanations for recommendation.
Extensive experiments show that \proposed outperforms existing methods on both rating prediction and candidate item reranking for top-\textit{k} recommendation, while significantly enhancing the explainability of recommendation systems.
Our code and data are available at {\url{https://github.com/jieyong99/EXP3RT}}.

\begin{figure}[t]
\centering
    \includegraphics[width=\columnwidth]{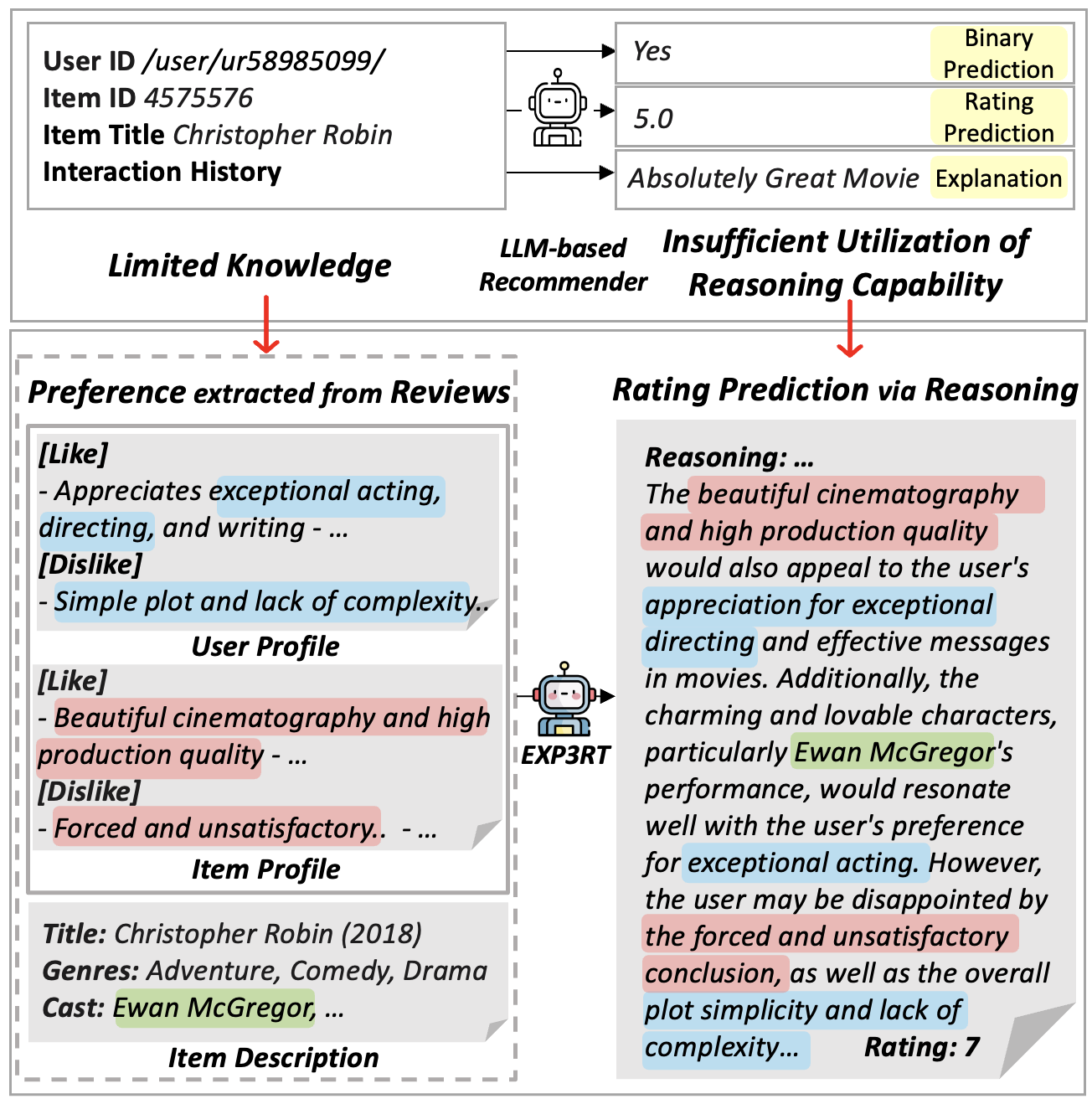}
    \caption{Comparison between existing LLM-based recommenders (Upper) and our proposed \proposed (Lower). \proposed predicts the rating score of a user-item pair based on step-by-step textual reasoning, by leveraging key preference descriptions extracted from user and item reviews.}
    \label{fig:intro} 
\end{figure}

\end{abstract}


\maketitle

\section{Introduction}
\label{sec:intro}


Large Language Models (LLMs) have demonstrated remarkable performance in diverse tasks attributed to their exceptional ability to understand semantic information and perform advanced reasoning~\cite{Brown2020LanguageMA,Kojima2022LargeLM,Wei2022ChainOT}.
Leveraging these strengths, there have been many attempts to integrate LLMs into recommender systems to enhance overall performance.
Some studies have explored in-context learning for recommendation tasks by directly prompting LLMs~\cite{Gao2023ChatRECTI,Liu2023IsCA};
however, this approach often struggles with limited accuracy due to insufficient domain-specific training.
As a result, recent studies mainly focus on fine-tuning LLMs to better adapt them to specific recommendation tasks and datasets. 
For instance, LLMs have been fine-tuned for binary prediction~\cite{Bao2023TALLRecAE, Zhang2023CoLLMIC}, user rating prediction~\cite{Geng2022RecommendationAL}, next item prediction~\cite{Ji2023GenRecLL}, and candidate item reranking~\cite{Yue2023LlamaRecTR}.

Despite recent advancements, existing studies partially capitalize on leveraging the potential of LLMs for recommendation tasks due to several limitations. 
First, they often rely on insufficient input information about users and items, such as user/item IDs, item titles, and interaction history~\cite{Geng2022RecommendationAL,Bao2023TALLRecAE,Yue2023LlamaRecTR,Zhang2023CoLLMIC,Ji2023GenRecLL}, which restricts the model's understanding of user preferences (Figure~\ref{fig:intro}, Upper Left). 
This limited information hinders the generalization capability of LLM-based recommenders in predicting outputs. 
Second, these models are typically optimized to generate short responses, such as simple ``yes''/``no'' answers or rating scores~\cite{Geng2022RecommendationAL,Bao2023TALLRecAE,Zhang2023CoLLMIC}, where the lack of rich target context makes it challenging for the model to produce accurate predictions (Figure~\ref{fig:intro}, Upper Right).
Besides, concise outputs fail to provide logical explainability for recommendations.

On the other hand, another line of research has focused on generating personalized explanations for recommendations by leveraging the text generation capabilities of LLMs~\cite{Geng2022RecommendationAL, Li2022PersonalizedPL, Ma2024XRecLL}. 
This involves first annotating an explanation for a specific user-item interaction using the corresponding review text and then optimizing the LLM to generate this annotated explanation;
for instance, some studies simply use the first sentence or a brief summary of the review as the target explanation~\cite{Geng2022RecommendationAL,Li2022PersonalizedPL,Ma2024XRecLL}. 
However, annotations taken directly from target reviews often fail to generalize to unobserved user-item interactions, due to their lack of grounding in evidence or logical reasoning.
Consequently, the generated explanations may exhibit poor faithfulness and reasonableness.
To address these limitations, our work aims to fine-tune an LLM for \textit{personalized preference reasoning}, enhancing both rating prediction accuracy and explainability by leveraging user and item review texts. 
Reviews contain rich details about users' subjective opinions, making them an excellent resource for modeling user and item preferences.
Our key idea is utilizing \textit{user profiles} and \textit{item profiles}, which refer to lists of key preference descriptions extracted from reviews (Figure~\ref{fig:intro}, Lower Left).
These profiles mitigate the impact of noise in raw review data and enable a semantic understanding of users’ and items’ preferences.
Using these profiles, we aim to conduct step-by-step textual reasoning that leverages structured user and item preferences to enhance rating prediction accuracy, while also providing faithful and logical explanations for recommendation.

A straightforward approach to performing step-by-step textual reasoning is to leverage high-performing yet expensive LLMs.  
However, such models face major challenges in domain-specific adaptation and incur high computational overhead.
To overcome these issues, we adopt a knowledge distillation framework~\cite{hinton2015distilling}; given an observed user-item rating, a teacher LLM (e.g., GPT-3.5) annotates high-quality textual reasoning aligned with the rating score, and a student LLM (e.g., LLaMA3-8B) is fine-tuned to first generate this reasoning and then predict the rating score.
Through this process, the student LLM inherits advanced reasoning capabilities for rating prediction from the teacher LLM, allowing it to better understand user-item interactions and preferences.
As a result, it achieves effective domain-specific adaptation while significantly reducing computational costs.



In this work, we introduce \textbf{\proposed}, \underline{\textbf{EXP}}lainable \underline{\textbf{P}}ersonalized \underline{\textbf{P}}reference \underline{\textbf{R}}easoner for recommenda\underline{\textbf{T}}ion, which performs reason\\ing-enhanced rating prediction based on review-driven preference profiles of users and items.
To be specific, \proposed is fine-tuned as a student LLM via distillation from a teacher LLM for three different steps, each of which correspond to each step: 
(1) \textit{preference extraction}, the first step which encapsulates preference sets by extracting them from raw reviews, (2) \textit{profile construction}, the second step which constructs user and item profiles by aggregating and summarizing the preference sets, and (3) \textit{reasoning-enhanced rating prediction}, the final step that predicts the user's rating through step-by-step textual reasoning.
This distillation process makes our student LLM cost-efficient as well as significantly improve its reasoning capability.
During inference, \proposed utilizes both subjective and objective information from user/item profiles and item descriptions, allowing it to comprehensively understand user and item preferences. 
This process leads to effective generalization to unobserved user-item interaction pairs.


Our extensive experiments evaluate the effectiveness of \proposed for various recommendation tasks, including rating prediction, top-\textit{k} recommendation, and explanation generation.
The evaluation results demonstrate that the personalized preference reasoning of \proposed not only improves rating prediction and reranking accuracy but also offers explainability both logical and easy to understand.
Additionally, \proposed effectively reranks candidate items, generated by a conventional Collaborative Filtering (CF) method, within a multi-stage ranking pipeline for top-$k$ recommendation.

The main contributions are summarized as follows:
\begin{itemize}
    \item \proposed effectively enhances rating prediction accuracy via review-driven personalized reasoning.
\end{itemize}
\begin{itemize}
    \item \proposed generates detailed step-by-step reasoning, providing faithful and logical explanations by utilizing rich preference information extracted from reviews.
\end{itemize}
\begin{itemize}
    \item \proposed can function independently as a recommender and flexibly integrate with traditional CF-based recommenders as an item reranker within a multi-stage ranking pipeline.
\end{itemize}

\section{Related Work}
\label{sec:relwork}

\begin{table*}[t]
\centering
\small
\caption{Comparison of LLM-based recommenders, categorized by their target tasks.
\textit{Rerank-ability} denotes whether each method can function as a candidate item reranker within a multi-stage ranking pipeline for top-\textit{k} recommendation.
\pfive~\cite{Geng2022RecommendationAL} achieves both rerank-ability and explainability via multi-task learning (i.e., explanation generation and rating prediction).}
\label{tbl:model_comparison}
\resizebox{0.99\linewidth}{!}{
\begin{tabular}{lllcccccccccc}
\toprule
{\textbf{Task}} & {\textbf{Method}} & {\textbf{LLM Input}} & \textbf{Rerank-ability} & \textbf{Explainability} \\
\midrule
Multiple Tasks$^*$ & \pfive~\cite{Geng2022RecommendationAL} & User/Item ID, Item Title, Feature Word & \cmark & Review Generation  \\ 
Explanation Generation & \pepler~\cite{Li2022PersonalizedPL}, \xrec~\cite{Ma2024XRecLL} & User/Item Embedding, User/Item Feature & \xmark & Review Generation   \\ 
Target Item Generation & \genrec~\cite{Ji2023GenRecLL},  \llara~\cite{liao2024llara}, \slim~\cite{wang2024slim}  & Interaction History (+User/Item Embedding) & \cmark & \xmark  \\ 
Candidate Item Reranking & \zsranker~\cite{Hou2023LargeLM}, \llamarec~\cite{Yue2023LlamaRecTR} & Interaction History, Item Candidates & \cmark & \xmark  \\
Binary Prediction & \tallrec~\cite{Bao2023TALLRecAE}, \collm~\cite{Zhang2023CoLLMIC}, \trsr~\cite{zheng2024harnessing} & Interaction History (+User/Item Embedding) & \cmark & \xmark  \\
{Rating Prediction} & \llmrec~\cite{Liu2023IsCA} & Interaction History, Item Title, Item Category & \cmark & \xmark  \\ \midrule
{Rating Prediction} & \textbf{\proposed} (ours) & User/Item Profile, Item Description & \cmark & Step-by-step Reasoning \\  
\bottomrule
\end{tabular}
}
\end{table*}

\subsection{LLM-based Recommenders for Prediction}
\label{subsec:direct_rec}

Recent advancements in LLMs have led to significant performance improvements across various tasks~\cite{Brown2020LanguageMA}, resulting in a growing number of attempts to harness their advanced capabilities in recommender systems (RS) \cite{Wu2023ASO}.
Some approaches have utilized LLMs to provide binary recommendations (i.e., \textit{yes} or \textit{no}) to a user for target items~\cite{Bao2023TALLRecAE, Zhang2023CoLLMIC, zheng2024harnessing} or to predict the user's ratings on the items~\cite{Geng2022RecommendationAL}.
For binary recommendations, \tallrec~\cite{Bao2023TALLRecAE} utilizes a two-stage instruction tuning process on Alpaca~\cite{taori2023stanford} to effectively handle few-shot scenarios.
Furthermore, \trsr~\cite{zheng2024harnessing} more effectively leverages the text comprehension capabilities of LLMs by recurrently summarizing user history and utilizing it for recommendations.
In a different approach, \collm~\cite{Zhang2023CoLLMIC} integrates collaborative filtering information directly into the LLM’s input embedding space, combining the strengths of traditional collaborative filtering methods with the advanced capabilities of LLMs for more informed predictions.

For rating predictions, both \pfive~\cite{Geng2022RecommendationAL} and \llmrec~\cite{Liu2023LLMRecBL} focus on predicting user ratings by utilizing user and item IDs, enabling the models to generate predictions based on these identifiers.
Other studies shift their focus towards item generation and re-ranking tasks. 
\genrec~\cite{Ji2023GenRecLL} applies instruction tuning on LLaMA~\cite{Touvron2023LLaMAOA} to create generative recommendations, adapting the model to generate relevant items for users.
\llara~\cite{liao2024llara} utilizes hybrid prompting to integrate ID-based embeddings and textual features, \slim~\cite{wang2024slim} employs step-by-step knowledge distillation for resource-efficient reasoning, and \allmrec~\cite{kim2024allmrec} leverages collaborative filtering knowledge from pre-trained recommender system models, all to utilize LLMs for sequential recommendation.
\llamarec~\cite{Yue2023LlamaRecTR} and Zero-Shot Ranker~\cite{Hou2023LargeLM} investigate the potential of LLMs in re-ranking candidate items that have been initially generated by a separate retrieval model, highlighting the adaptability of LLMs in refining recommendation lists.
Additionally, recent studies have focused on utilizing user history to construct textual profiles with LLMs for performing top-k recommendations via traditional recommenders~\cite{wei2024llmrec, ren2024representation}.
A comparison of these methods can be found in Table \ref{tbl:model_comparison}.


Despite these efforts, the aforementioned studies have achieved limited accuracy in preference prediction tasks compared to conventional CF methods;
they fail to fully utilize exceptional reasoning abilities of LLMs.
First, they rely on limited contextual information about users and items, which is insufficient to make LLMs understand the latent preference of a target user and item.
Most of them take only user/item IDs or minimal attributes as inputs.
In addition, they have not explored advanced textual reasoning strategy, such as Chain-of-Thought (CoT) reasoning~\cite{Kojima2022LargeLM}, in that their models are simply optimized to generate short-form outputs tailored for their target tasks.
In this sense, our work aims to leverage rich information sources, specifically \textit{reviews}, and to predict user ratings through enhanced reasoning, supported by distillation methods to fully harness the advanced reasoning capabilities of LLMs.

\subsection{LLM-based Recommenders for Explanation}
\label{subsec:review_based_rec}
LLMs are highly effective at utilizing and processing semantic information to generate coherent and relevant text.
Consequently, some studies have focused on leveraging this capability by using sources rich in semantic content, such as reviews~\cite{Geng2022RecommendationAL,Ma2024XRecLL}, to generate explanations or summarize reviews for recommendations.
For explanation generation, \pfive~\cite{Geng2022RecommendationAL} uses user and item IDs combined with feature words to guide the generation of explanations.
\pepler~\cite{Li2022PersonalizedPL} integrates user and item ID vectors within a pretrained transformer, applying sequential tuning and recommendation regularization to align prompts with the model’s capabilities.
Meanwhile, \xrec~\cite{Ma2024XRecLL} leverages collaborative signals along with a lightweight collaborative adaptor, which helps LLMs better grasp complex user-item interaction patterns and generate more accurate explanations.
For review summarization, \pfive uses user IDs alongside the reviews themselves, allowing for the creation of summaries that reflect the specific perspectives and experiences of individual users.


Despite their effectiveness, these studies have notable limitations.
They often focus on merely extracting sentences or generating summaries from a single review, which can result in poor generalization and a lack of grounding in evidence or logical reasoning, making it difficult for these explanations to adapt to unobserved user-item interactions.
To overcome these limitations, our work focuses on extracting key information of user and item preferences from reviews and encapsulating them into a structured format. We further aggregate and summarize these insights from multiple reviews to build well-structured collective knowledge, enabling a comprehensive understanding of user and item preferences.

\section{Proposed Method: \proposed}
\label{sec:method}

\begin{figure*}[!thbp]
    \centerline{\includegraphics[width=\textwidth]{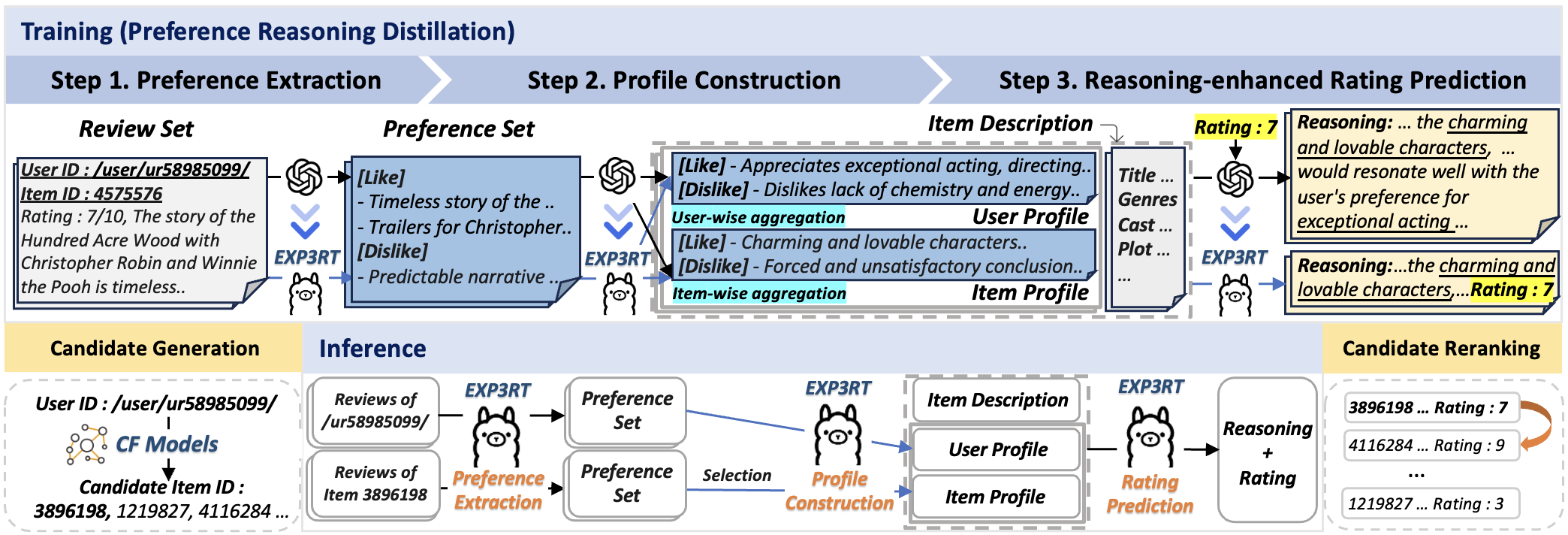}}
    \caption{
    The overview of our \proposed framework. 
    During training, we distill the reasoning capabilities of a teacher LLM (i.e., \gptthree) into our student LLM (i.e., Llama3-8B) for three steps:
    (1) Extracting preference descriptions from raw reviews,
    (2) constructing user/item profiles by aggregating these preferences, and
    (3) predicting rating scores based on textual reasoning.
    During inference, given a user-item pair, \proposed sequentially performs these steps to predict the rating;
    in addition, it can serve as an item reranker for top-\textit{k} recommendation, compatible with other CF models that efficiently retrieve candidate items.}
    \label{fig_method} 
\end{figure*}

In this section, we present \proposed, an LLM-based recommendation model designed to enhance both accuracy and explainability by leveraging subjective user and item preferences extracted from reviews, as well as objective item descriptions. 
\proposed is optimized by distilling the remarkable reasoning capabilities of a teacher LLM into a student LLM, for three key steps:
(1) \textit{preference extraction},
(2) \textit{profile construction}, and
(3) \textit{textual reasoning for rating prediction}.

Given an input user-item pair, \proposed first extracts key preference information from each review related to the user and item, organizing this information into a list. 
Next, \proposed constructs profiles for the user and item by aggregating and summarizing all extracted preferences in a user-wise and item-wise manner, respectively.
Finally, \proposed predicts the user's rating on the item through step-by-step textual reasoning, using the user and item profiles along with the item description as inputs.
Figure~\ref{fig_method} illustrates the overview of our \proposed framework.

\subsection{Preliminaries}
\subsubsection{Rating prediction}
Given a user set \userset and an item set \itemset, we collect a set of user-item interactions, which includes ratings (denoted by \interactions) and reviews (denoted by \reviews) written by users to items.
The problem of rating prediction involves estimating the rating \actualrating for a user-item pair \useritempair based on observed ratings \interactions and reviews \reviews. 
That is, it aims to accurately predict the rating \predrating$\approx$\actualrating for any unobserved user-item pair \useritempair, where \predrating falls within the predefined rating scale $S$.

\subsubsection{Knowledge distillation from teacher to student LLM}
LLMs like \gptthree demonstrate remarkable reasoning abilities but struggle with recommendation tasks when primarily relying on in-context learning.
This work leverages the strengths of such models through distillation into a more cost-efficient one, effectively utilizing their key capabilities while optimizing computational efficiency.
Specifically, we use GPT-
3.5\footnote{\url{https://chat.openai.com/}} as the teacher LLM, leveraging its superior reasoning and language understanding capabilities. 
The student LLM is built upon LLaMA3-8B\footnote{\url{https://llama.meta.com/docs/model-cards-and-prompt-formats/meta-llama-3}} model, of which capabilities would be enhanced through distillation from the teacher LLM.
The knowledge distillation process primarily consists of two steps: (1) constructing training data by prompting the teacher LLM to generate high-quality outputs, and (2) optimizing the student LLM through supervised fine-tuning (SFT) using the generated data, specifically designed to guide the student LLM in mimicking the teacher LLM’s outputs.
This approach enables the student LLM to perform effectively on various tasks while reducing resource requirements.


\subsection{Preference Extraction from Reviews}
\label{subsec:pref_extract}
The first step of \proposed is to extract key preferences from raw reviews and organize them in a structured format.
Since raw reviews may be unstructured and noisy, making it difficult for LLMs to understand the contents, we extract and encapsulate a review into a list of preference descriptions, named as \textit{preference}.
Specifically, the preference categorizes a user's various tastes for an item into ``{\textit{Like}}'' and ``{\textit{Dislike}}'' which can be inferred from the review.

To distill the teacher LLM's preference extraction ability, we collect training examples by prompting the teacher LLM to extract key preferences (i.e., \textit{Like} and \textit{Dislike}) from an input review $v$ and organize them into a list of preference descriptions $p$, i.e.,
${D}_{\text{PE}}=\{(v, p)| p\sim P_\text{teacher}(\cdot|v,T_{\text{PE}}), v\in\mathcal{V}\}$ where $D_{\text{PE}}$ is the dataset, $T_{\text{PE}}$ is the prompt for the preference extraction step, and $P_\text{teacher}$ is the decoding probability of the teacher LLM.

\subsection{User and Item Profile Construction}
\label{subsec:prof_const}
The second step is constructing \textit{user and item profiles} $s$, which are texts reflecting comprehensive subjective preference information about users or items; 
the profiles are created by aggregating and summarizing preferences extracted from all \reviews (in Section~\ref{subsec:pref_extract}) into a structured format for each user or item. 
Utilizing these profiles reduces noise in reviews and captures key preferences and characteristics, enabling more effective reasoning in recommendation.

The user profile $s_u$ is constructed by aggregating the set of preferences $P_u=\{p|(v\in\mathcal{V}_u,p)\in D_\text{PE}\}$ extracted from the user’s review set $\mathcal{V}_u$, 
while the item profile $s_i$ is created in a same way by using the set of preferences $P_i=\{p|(v\in\mathcal{V}_i,p)\in D_\text{PE}\}$ obtained from the item’s review set $\mathcal{V}_i$. 
It can also be constructed using preferences chosen with specific criteria (in Section~\ref{subsec:inference}).
To achieve this, training examples for profile construction are obtained by prompting the teacher LLM, i.e., ${D}_{\text{PC}}=\{(P, s)| s\sim P_\text{teacher}(\cdot|P,T_{\text{PC}})\}$, where $T_{\text{PC}}$ is the prompt for the profile construction step.

\subsection{Reasoning-enhanced Rating Prediction}
\label{personalized_reasoning} 
The final step is to predict the rating \predrating of a user-item interaction based on step-by-step textual reasoning, utilizing key information extracted from \reviews (in Sections~\ref{subsec:pref_extract} and \ref{subsec:prof_const}).
Since user/item profiles contain subjective information, we additionally make use of item descriptions as an objective information source, in order to generate reasoning from a comprehensive perspective.
In the reasoning process, our model is expected to match user preferences (from user profile) with item characteristics (from item profile and descriptions) to evaluate their alignment for rating prediction.
Consequently, through step-by-step reasoning, \proposed generates personalized and comprehensible reasonings, which are helpful to predict a user's rating for a target item and provide logical explanation for supporting the recommendation.



To achieve this, we annotate detailed step-by-step reasoning for each user-item interaction (i.e., rating) by leveraging the great reasoning capability of the teacher LLM.
For an observed interaction between user $u$ and item $i$, we prompt the teacher LLM to generate a textual explanation $e$ given a user profile $s_u$, an item profile $s_i$, item descriptions $d_i$, and the user's actual rating $r_{u,i}$;
that is, the training dataset for reasoning-enhanced rating prediction $D_{\text{RP}}$ is obtained by ${D}_{\text{RP}} = \{(s_u, s_i, d_i, e, r_{u,i})|e\sim P_\text{teacher}(\cdot|s_u, s_i, d_i, r_{u,i}, T_{\text{RG}}) \},$
where $T_{\text{RG}}$ is the prompt for the reasoning generation step.


\subsection{Optimization}
\label{tuning_method}
Instruction tuning is a key technique for training LLMs with human-annotated instructions, enabling them to handle tasks more effectively~\cite{Ouyang2022TrainingLM}.
To apply LLMs to recommendation tasks, we use this technique to guide the model in predicting user preferences.
To perform instruction tuning, we transform the three datasets ${D}_{\text{PE}}$, ${D}_{\text{PC}}$, and ${D}_{\text{RP}}$ into the prompts to include specific instructions tailored for their steps.
Then, we combine them to obtain the final dataset $\mathcal{D}$, which consists of instruction input-output pairs $(x,y)$ for the three steps.
We employ QLoRA~\cite{Dettmers2023QLoRAEF} for each step, making our student LLM efficiently adapt to various tasks.
Note that QLoRA enables us to integrate supplementary information while keeping the original parameters unchanged.

The final learning objective of \proposed is defined by Negative Log-Likelihood (NLL), used to maximize the probability of the correct token \( y_t \) given the previous tokens \( y_{<t} \) and the input \( x \) as follows:
\begin{equation}
\max_{\Theta} \sum_{(x,y) \in \mathcal{D}} \sum_{t=1}^{|y|} \log \left( P_{\Phi+\Theta} \left( y_t \mid x, y_{<t} \right) \right),
\end{equation}
where \(\Theta\) represents the QLoRA parameters, and \(\Phi\) represents the parameters of our target student LLM, and only the QLoRA parameters are updated during the training process.

\subsection{Inference}
\label{subsec:inference}
For inference on an unobserved user-item $(u, i)$ pair, \proposed first extracts information that reflects preferences from their raw reviews, $\mathcal{V}_u$ and $\mathcal{V}_i$, to form preference sets, $P_u$ and $P_i$.
Then, it constructs the user/item profile, $s_u$ and $s_i$, by aggregating and summarizing the preference sets, following the processes described in Sections~\ref{subsec:pref_extract} and~\ref{subsec:prof_const} to ensure consistency.
If a profile for the user or item has already been created, it is directly used in the inference process. 

For user profile construction, we utilize all available reviews as it is crucial to gather as much information as possible about the user to ensure accurate and personalized recommendations.
In contrast, item profile can benefit from a selective aggregation approach, focusing on relevant attributes that directly influence user preferences.
In detail, the set of item preferences can be determined in three different ways:
(1) using all preferences of the item to reflect the overall preferences for the item,
(2) selecting only the preferences extracted from the reviews with the highest number of helpful votes (i.e., \textit{helpfulness-based selection}) to highlight the item's representative preferences, and
(3) using the preferences from the reviews written by the users with similar preferences to a target user (i.e., \textit{neighbor-based selection})~\cite{liu2014new} to create a more personalized item profile.\footnote{The details of calculating user preference similarity can be found in Appendix~\ref{sec:item_synopsis_construction}.}
The preferences selected through these various criteria are then used to construct the profiles, following the process described in Section~\ref{subsec:prof_const}.
By default, we construct item profiles using all preferences of the item. 

Finally, \proposed takes the user profile, item profile, and item descriptions as its input, to predict the rating score along with textual reasoning.
In the input prompt for \proposed, the average ratings for both the user and the item are included, which represent biases to address label inconsistency.
Through detailed textual reasoning that evaluates matches and mismatches between preferences of the user and item, \proposed generates comprehensive explanation and predicts the user’s rating for the target item.

\subsubsection{Rating Score Prediction}
\label{det_rat_pred}
To output the rating scores as continuous real values for more precise prediction beyond the limitations of discrete integer values, we adopt a post-computation technique by utilizing the model's logit outputs for the numeric tokens.
Training the model to predict ratings as discrete categories (i.e., integer tokens) can reduce the precision of predictions. 
For example, items predicted to have a rating of 3 might actually be closer to 2 or 4.
Such a decoding approach can mask significant differences between items, as assigning them the same integer value fails to capture subtle variations in their actual ratings.

Given the logit values $l_i$ corresponding to the model's tokens for integer rating $r_i$, we obtain the probability $p(r_i)$ of each rating by using the softmax function, $p(r_i) = {\exp({l_i})}/{\sum_{j=1}^{|S|} \exp({l_j})}$.
We use the expected value of integer ratings as our final prediction:
\begin{equation}
\mathbb{E}[R] = \sum_{r_i} r_i \cdot p(r_i).
\end{equation}
This method enables continuous rating predictions, even though the model is trained to decode ratings as discrete integers.

\subsubsection{Top-\textit{k} Recommendation with \proposed}
\label{subec:pipeline}
Since \proposed is built upon an LLM with billion-scale parameters, directly using our model for top-\textit{k} recommendation incurs significant computational costs and is practically infeasible.
On the other hand, traditional CF-based recommenders achieve decent performance with relatively lower costs when generating candidate items for top-\textit{k} recommendation.
We incorporate \proposed into a multi-stage ranking pipeline~\cite{covington2016deep}, where traditional CF-based recommendation models serve as candidate generators.
After efficiently generating these candidates, \proposed reranks them based on their predicted ratings to produce a final ranked list of the top-\textit{k} items.
This approach allows our LLM-based recommender to effectively handle top-\textit{k} recommendation while maintaining relatively low computational costs.
In short, the multi-stage ranking pipeline combines the scalability and efficiency of CF-based candidate generation with the accuracy of LLM-based reranking, making it highly adaptable for a variety of recommendation scenarios.

\section{Experiments}
\label{sec:exp}
In this section, we design and conduct experiments\footnote{We use a dedicated Advanced Database System Infrastructure (NFEC-2024-11-300458).} to answer the following research questions:



\begin{itemize}[leftmargin=*,topsep=2pt,itemsep=2pt,parsep=0pt]
    \item \textbf{RQ1:} How effectively does \proposed perform in rating prediction, compared to existing methods?

    \item \textbf{RQ2:} How effectively does \proposed rerank candidate items within the multi-stage ranking pipeline for top-\textit{k} recommendation?

    \item \textbf{RQ3:} How good personalized reasoning can \proposed generate?
    

    \item \textbf{RQ4:} How effectively do the components and configurations of \proposed contribute to its overall performance?
\end{itemize}
\begin{table*}[t]
\centering
\caption{Performance for rating prediction. The best and second-best results are in bold and underlined (*: $p$-value < 0.05).}
\label{tbl:rating_prediction_table}
\resizebox{0.99\linewidth}{!}{
\begin{tabular}{lSSSSSSSSSS}
\toprule
\multirow{3.5}{*}{\textbf{Methods}} & \multicolumn{2}{c}{\multirow{2.5}{*}{\textbf{\imdb}}} & \multicolumn{8}{c}{\textbf{\amazon}} \\ \cmidrule(lr){4-11}
& & & \multicolumn{2}{c}{\textbf{Total}}  & \multicolumn{2}{c}{\textbf{Warm-start}}  & \multicolumn{2}{c}{\textbf{Cold-start}}  & \multicolumn{2}{c}{\textbf{Unseen}} \\
\cmidrule(lr){2-3}\cmidrule(lr){4-5}\cmidrule(lr){6-7}\cmidrule(lr){8-9}\cmidrule(lr){10-11}
 & \textbf{RMSE} & \textbf{MAE} & \textbf{RMSE} & \textbf{MAE} & \textbf{RMSE} & \textbf{MAE} & \textbf{RMSE} & \textbf{MAE} & \textbf{RMSE} & \textbf{MAE} \\ \midrule

\mf & \underline{1.9530} & \underline{1.4771} &  \underline{0.6683} & 0.4572 & \underline{0.6663} & \underline{0.4588} & \underline{0.6769} & 0.4501 & - & - \\
\mlp & 2.2562 & 1.7467 & 0.7020 & 0.5324  & 0.6963 & 0.5280 & 0.7262 & 0.5519 & - & - \\
\rgcl & 2.3143 & 1.8036  & 0.7562 & 0.6117 & 0.7533 & 0.6109 & 0.7693 & 0.6154 & - & - \\ \midrule
\pfive & 2.5471 & 1.8604 & 0.8868 & 0.4756 & 0.8873 & 0.4810 & 0.8846 & 0.4508 & - & - \\
\llmrec (\gptthree, ZS) & 2.4843 & 1.7559 & 0.8974 &  0.6356 & 0.8910 & 0.6301 & 0.9260 & 0.6609 & 1.9732 & 1.6170  \\
\llmrec (\gptthree, FS) & 2.2161 & 1.6223 & 0.7852 & \underline{0.4562} & 0.7888 & 0.4623 & 0.7688 & \underline{0.4286} & 1.7607 & 1.3000 \\
\llmrec (\gptfour, ZS) & 2.1586 & 1.6261 & 0.9986 & 0.8331 & 0.9941 & 0.8257 & 1.0187 & 0.8670 & 1.7139 & 1.3958 \\
\llmrec (\gptfour, FS) & 2.1410 & 1.6334 & 0.8386 & 0.5004 & 0.8370 & 0.5010 & 0.8460 & 0.4974 & \underline{1.4993} & \textbf{1.0300}* \\
\proposed & \textbf{1.9122}* & \textbf{1.4691}* & \textbf{0.6508}* & \textbf{0.4297}* & \textbf{0.6572}* & \textbf{0.4369}*  & \textbf{0.6072}* & \textbf{0.3858}* & \textbf{1.4398}* & \underline{1.1192} \\ \bottomrule

\end{tabular}
}
\end{table*}

\subsection{Experimental Settings}
\label{subsec:exp_settings}
\subsubsection{Datasets}

We conduct our experiments on two datasets, \textbf{\imdb} \cite{kim-etal-2024-pearl} and \textbf{\amazon}~\cite{hou2024bridging}, to demonstrate the ability of \proposed to handle diverse review styles and content. 
\imdb
is a movie-domain dataset with ratings on a scale of 1 to 10, while \amazon
uses a scale of 1 to 5 and is widely used in prior research~\cite{Zhang2023CoLLMIC}. 
Both datasets include item metadata and user reviews with ratings.

Considering the significant computational cost associated with generating reasoning-enhanced explanations using \gptthree
, we strategically sample approximately 100k interactions, following the standard in LLM-based recommendation studies~\cite{Bao2023TALLRecAE, Yue2023LlamaRecTR, wang2024slim}.
To evaluate the robustness of \proposed in the rating prediction task, we configure varying dataset density levels to address a broad range of scenarios. 
Specifically, {\imdb} is configured with higher density compared to {\amazon}, which typically favors CF methods.
In contrast, {\amazon} is set with lower density to encompass a broader range of scenarios including cold-start, warm-start, and unseen cases.
For preprocessing, we exclude users without user ratings and apply a 5-core filtering strategy, iteratively removing users and items with fewer than five interactions.
As a result, the processed datasets comprise approximately 859 users and 1,155 items with 19,124 interactions for \imdb, and about 10,761 users and 10,084 items with 118,705 interactions for \amazon. 
The datasets are split into training, validation, and testing sets in an 8:1:1 ratio based on interaction timestamps~\cite{Zhang2023CoLLMIC}, simulating real-world recommendation scenarios and prevent data leakage. 
\subsubsection{Baselines}
We consider a range of baseline methods for three tasks: 
(1) rating prediction, (2) candidate item reranking, and (3) explanation generation.
We exclude models that do not perform each task directly or use LLMs in auxiliary roles~\cite{Zhang2023CoLLMIC,Ji2023GenRecLL,Yue2023LlamaRecTR,zheng2024harnessing,liao2024llara,wang2024slim}, to evaluate LLMs' recommendation capabilities. 

\smallsection{Rating prediction}
We consider a variety of baseline methods for rating prediction, including conventional CF methods~\cite{Koren2009MatrixFT,Cheng2016WideD}, a review-enhanced CF method~\cite{Shuai2022ARG}, and LLM-based methods~\cite{Geng2022RecommendationAL,Liu2023IsCA}.
\begin{itemize}
[leftmargin=*,topsep=2pt,itemsep=2pt,parsep=0pt]
    \item \textbf{\mf}~\cite{Koren2009MatrixFT}: Matrix Factorization, which is one representative latent factor-based CF method.
    \item \textbf{\mlp}~\cite{Cheng2016WideD}: Wide and Deep Learning, jointly training wide linear models and deep neural networks, to combine the benefits of memorization and generalization.
    \item \textbf{\rgcl}~\cite{Shuai2022ARG}: Review-aware Graph Contrastive Learning framework for review-based recommendation.
    \item \textbf{\pfive}~\cite{Geng2022RecommendationAL}: A text-to-text LLM for recommendation, which incorporates various recommendation tasks into a shared framework. 
    We employ T5-base~\cite{raffel2020exploring} as the backbone LLM.
    \item \textbf{\llmrec}~\cite{Liu2023IsCA}: A prompting-based method using an LLM as a recommender in rating prediction. We adopt zero-shot (ZS) and few-shot (FS) prompting with \gptthree and \gptfour.
\end{itemize}

\smallsection{Candidate item reranking}
We explore personalized item ranking capability of existing LLM-based methods specifically designed for preference prediction~\cite{Bao2023TALLRecAE,Hou2023LargeLM}. 
\begin{itemize}
[leftmargin=*,topsep=2pt,itemsep=2pt,parsep=0pt]
    \item \textbf{\tallrec}~\cite{Bao2023TALLRecAE}: Lightweight instruction tuning framework which generates a binary answer of ``Yes'' or ``No'' by composing a task instruction based on their historical interactions. We rerank the items with the LLM's output logit value.
    \item \textbf{\zsranker}~\cite{Hou2023LargeLM}: A prompting-based method that uses an LLM as a zero-shot ranker, based on sequential interaction histories and retrieved candidate items. We employ \gptthree as the LLM.
\end{itemize}
      
\smallsection{Explanation generation}
The third category generates explanations about the recommended items, by leveraging LLMs' text understanding and generation capabilities. 
We compare these methods to highlight the differences and strengths of our explanation.
\begin{itemize}
[leftmargin=*,topsep=2pt,itemsep=2pt,parsep=0pt]
    \item \textbf{\pfive}~\cite{Geng2022RecommendationAL}: A text-to-text LLM optimized for explanation generation, which generates random spans extracted from an actual review. The target explanation is the first sentence of the review. 
    We employ the same LLM used in the rating prediction task.
    \item \textbf{\xrec}~\cite{Ma2024XRecLL}: 
    An LLM-based method for explanation, optimized to generate a concise summary of a user's actual review. The summary of a single review is provided as the target explanation. 
    We use \gptthree as the LLM to adopt the method.
\end{itemize}

\begin{table*}[t]
\centering
\caption{Performance comparison of various LLM-based recommenders as candidate item rerankers for top-$k$ recommendation, Dataset: \imdb. The best and second-best results are highlighted in bold and underlined (*: $p$-value < 0.05).}
\resizebox{1\linewidth}{!}
{
\begin{tabular}{clcccccccc}
\toprule
\textbf{{Generator}} & \textbf{{Reranker}}  & {\textbf{Recall@5}} & {\textbf{NDCG@5}} & {\textbf{Recall@10}} & {\textbf{NDCG@10}} & {\textbf{Recall@15}} & {\textbf{NDCG@15}} & {\textbf{Recall@20}} & {\textbf{NDCG@20}}\\
\midrule 
\multirow{5}{*}{\bprmf} 
& -  & 0.0296 & 0.0328 & 0.0468 & 0.0379 & 0.0659 & 0.0450 & \multirow{2}{*}{$\vert$} & 0.0516\\ 
& \pfive & 0.0232 & 0.0229 & 0.0418 & 0.0306  & 0.0609 & 0.0383 & & 0.0467\\ 
& \tallrec & 0.0172 & 0.0192 & 0.0394 & 0.0282  & 0.0642 & 0.0379 & 0.0843 & 0.0448\\ 
& \zsranker & \underline{0.0392} & \textbf{0.0435}* & \underline{0.0592} & \underline{0.0482}  & \underline{0.0711} & \underline{0.0530} & \multirow{2}{*}{$\vert$} & \underline{0.0568}\\ 
& \proposed & \textbf{0.0395} & \underline{0.0419} & \textbf{0.0624}* & \textbf{0.0491}* & \textbf{0.0768}* & \textbf{0.0543}*  & & \textbf{0.0569}\\ 
\midrule
\multirow{5}{*}{\lightgcn} 
& - & 0.0401 & 0.0463 & 0.0693 & 0.0560 & \underline{0.0972} & 0.0664 & \multirow{2}{*}{$\vert$} & 0.0739\\ 
& \pfive & 0.0362 & 0.0375 & 0.0578 & 0.0456  & 0.0872 & 0.0571 & & 0.0682\\ 
& \tallrec & 0.0274 & 0.0294 & 0.0583 & 0.0423  & 0.0890 & 0.0543 & 0.1169 & 0.0650\\ 
& \zsranker & \underline{0.0461} & \underline{0.0520} & \underline{0.0717} & \underline{0.0593} & 0.0953 & \underline{0.0685} & \multirow{2}{*}{$\vert$} & \underline{0.0765}\\ 
& \proposed & \textbf{0.0543}* & \textbf{0.0560}* & \textbf{0.0846}* & \textbf{0.0669}* & \textbf{0.1006}* & \textbf{0.0734}*  & & \textbf{0.0796}\\ 
\bottomrule
\end{tabular}
}
\label{tbl:topktable}
\end{table*}
\subsubsection{Evaluation Metrics}
We use different evaluation metrics for each target task.
For rating prediction, we use Mean Absolute Error (MAE) and Root Mean Squared Error (RMSE). 
For top-$k$ recommendation, we compare ranking performance using Recall and Normalized Discounted Cumulative Gain (NDCG). 

\subsubsection{Implementation Details}
We employ \gptthree as a teacher LLM and Llama3-8B-Instruct as a student LLM.
While GPT-4 offers advanced capabilities, we choose GPT-3.5 as the teacher LLM for its ability to generate high-quality explanation at a substantially lower computational cost, ensuring a balance between efficiency and effectiveness;
the teacher LLM is used to construct training data for \proposed.
User profiles, item profiles, and average ratings are constructed solely based on the samples in the training dataset.
All prompts used for constructing dataset are provided in the Appendix~\ref{sec:prompt}.
During training, we compute the loss only for label tokens within the prompt for each data instance.
This approach is more efficient, as optimizing the entire input does not yield significant improvements.
We conduct each experiment three times using different random seeds, and report the average values.
The models are trained with the AdamW optimizer using a learning rate of 2e-4, with a constant learning rate scheduler, and a weight decay of 0.0.
The maximum epoch is set to 10, with early stopping based on evaluation loss after each epoch and a patience of 1.
The maximum token length is set to 1200.
The model with the best validation performance is saved for testing set evaluation.
The QLoRA hyperparameters differ by dataset: for {\imdb}, we use lora\_r=512, lora\_alpha=128, and lora\_dropout=0.1; for {\amazon}, we use lora\_r=128, lora\_alpha=32, and lora\_dropout=0.1.
Further details about baselines are provided in the Appendix~\ref{sec:implementation_detail}.



\subsection{Rating Prediction Performance (RQ1)}
\label{subsec:rq1}

To validate the effectiveness of \proposed in the rating prediction task, we evaluate \proposed and other methods on both datasets.
Table~\ref{tbl:rating_prediction_table} shows that \proposed outperforms all other methods, consistently achieving superior results across both datasets.
Notably, even when compared to MF, which is recognized for its strong performance across domains among CF-based models, \proposed demonstrates higher efficacy.
When evaluated against LLM-based baselines, \proposed shows greater performance improvements. 
This result indicates that our model effectively leverages the reasoning capabilities of LLMs, achieving significant enhancements in rating prediction accuracy. 
The consistent outperformance of \proposed across various baselines validates its design and underlying approach.

To further assess the robustness of \proposed in various recommendation scenarios, we have designed settings for \textit{warm-start}, \textit{cold-start}, and \textit{unseen} user/item interactions:\footnote{We focus on a sparse dataset (i.e., \amazon) to identify cold-start users/items.} 
We divide the testing set into warm-start and cold-start subsets:
(1) The warm-start subset includes users who have interacted with items more than three times in the training set, (2) the cold-start subset comprises the remaining interactions, and (3) the unseen set consists of interactions from users who have not appeared in the training set, which is sampled from the original \amazon dataset.


In Table~\ref{tbl:rating_prediction_table}, we observe performance enhancements in both warm-start and cold-start scenarios for \proposed.
However, the performance gap between \proposed and the baselines is notably larger in the cold-start subset when compared to the original testing set.
This highlights that even in challenging scenarios like the cold-start problem, where the number of reviews available for constructing a profile is highly limited, our method can still construct profiles and perform accurate rating predictions without substantial limitations.
Furthermore, existing CF methods, which fall into a transductive approach, cannot predict the rating for unseen users and items, as they are not included in the training set.
On the contrary, \proposed is able to make predictions as long as user (or item) reviews are available, making it highly adaptable to real-world environments.

\begin{figure}[h]
\centering
    \includegraphics[width=\linewidth]{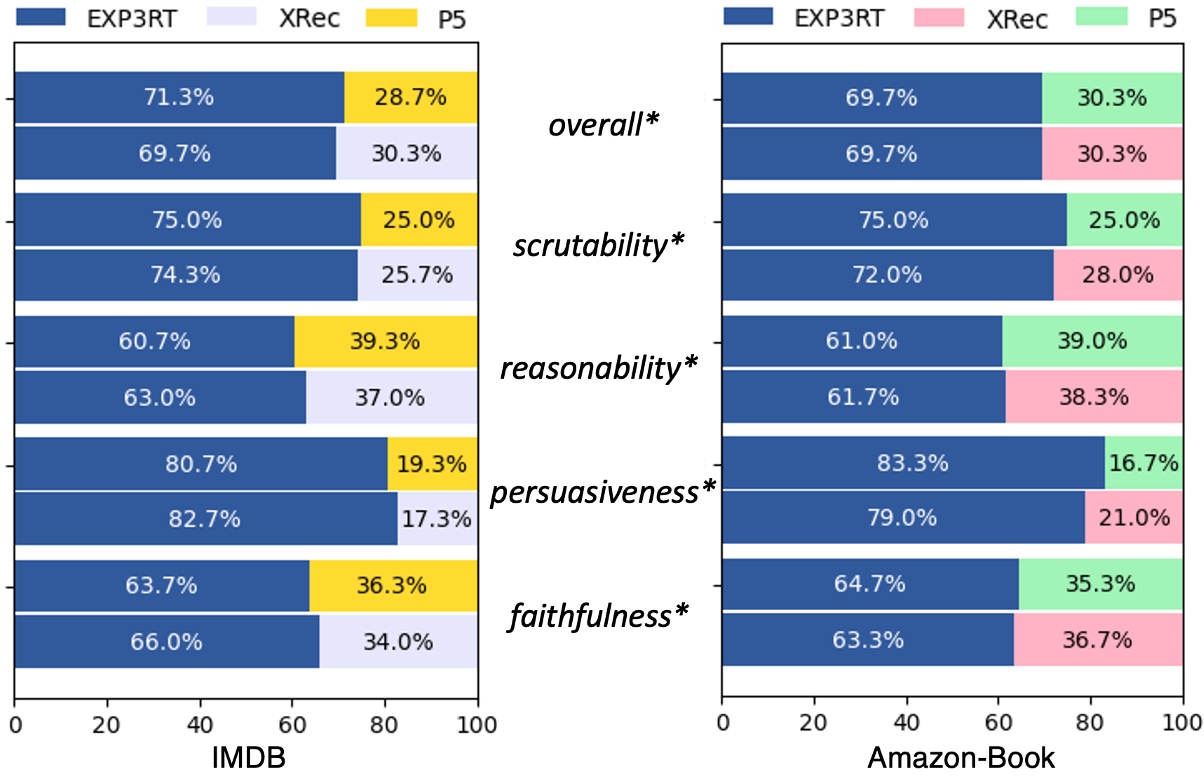}
    \caption{Human evaluation on pairwise quality comparison of recommendation explanations, generated by \proposed and baselines (*: $p$-value < 0.05). }
    \label{fig:humaneval} 
\end{figure}

\subsection{Top-\textit{k} Recommendation Performance (RQ2)}
\label{subsec:rq2}
To investigate the effectiveness of \proposed as a candidate item reranker within a multi-stage ranking pipeline, we evaluate the top-\textit{k} recommendation accuracy on the \imdb dataset.\footnote{The necessity of reranking top-\textit{k} candidate items, generated from a relevance-based recommender, becomes significant only for datasets having a balanced rating distribution, as the candidates need to be further aligned with users' true preference. Thus, we do not use \amazon having a skewed rating distribution (almost 5).}
To evaluate the model's ability to rank highly preferred items at the top, we create a balanced subset of the \imdb dataset by selecting interactions with ratings of 6 or higher from the original dataset.
Among these, items with ratings of 9 and 10 are clearly classified as highly preferred items and evenly split between the training and testing sets.
Through this process, the testing set contains only the most highly preferred items.
This setup allows us to test the reranker's ability to accurately identify and prioritize the most preferred items.

We choose two popular CF-based top-\textit{k} recommender, \textbf{\bprmf}~\cite{Koren2009MatrixFT} and \textbf{\lightgcn}~\cite{He2020LightGCNSA}, as candidate item generators to retrieve top-20 items for each user in the dataset.
As the rerankers, \proposed and other baseline methods rerank the retrieved candidate items based on the predicted scores or relevance.

Table~\ref{tbl:topktable} shows that \proposed considerably improves the recommendation accuracy by effectively reranking the candidate items.
In case of other LLM-based reranker (i.e., \pfive, \tallrec, and \zsranker), they sometime degrades the ranking accuracy compared to the case without the reranking process, whereas the reranking of \proposed consistently yields improvements.
These findings show the superiority of \proposed in discriminating between highly preferred and less preferred items within a subtle environment.

\subsection{Explanation Quality (RQ3)}
\label{subsec:rq3}
\begin{table}[t]
\centering
\caption{Explanations for user \textit{{AGIR2FRRAWXR6C4J7MZT5C QENTIA}} and item \textit{{B094NRQWQK}} (\textit{{Magnus Maximus 29: The Last Romans}}) generated by LLM-based methods.}
\begin{tabular}{P}
\toprule












\textbf{\pfive}: Max surprised me.
\\\midrule 
\textbf{\xrec}: The user would enjoy the item because it provides surprising confrontations and engaging interactions, which can add excitement and unpredictability to their experience.
\\\midrule
\textbf{\proposed}: The user would definitely enjoy ``Magnus Maximus 29: The Last Romans'' as it offers an epic storyline with entertaining twists and impressive performances. \textbf{The well-portrayed characters and action-packed plot would captivate the user's interest, and the cool twists in the story would keep user engaged.} Additionally, the increased stature of Max would likely appeal to the user's preference for impressive character development. Overall, the book seems to align perfectly with the user's love for killer stories and well-researched plots.
\\ \bottomrule
\end{tabular}
\label{tbl:casestudy}
\end{table}
\begin{table}[t]
\centering
\caption{Performance of \proposed for rating prediction with various item profile construction strategies (*: $p$-value < 0.05).}
\label{tbl:item_synopsis_rating_prediction_table}
\resizebox{0.99\linewidth}{!}{
\begin{tabular}{cUUUU}
\toprule
\multirow{2.5}{*}{\textbf{Selection Strategy}} & \multicolumn{2}{c}{\textbf{IMDB}} & \multicolumn{2}{c}{\textbf{Amazon-Book}} \\ 
\cmidrule(lr){2-3}\cmidrule(lr){4-5}
 & \textbf{RMSE} & \textbf{MAE} & \textbf{RMSE} & \textbf{MAE} \\ \midrule

All (\textit{Default}) & \textbf{1.9122}* & \underline{1.4691} & \textbf{0.6508}* & \textbf{0.4297}* \\ 
Helpfulness-based & \underline{1.9451} & \textbf{1.4530}* & 0.9727 & 0.7058 \\ 
Neighbor-based & 1.9812 & 1.4931 & \underline{0.9555} & \underline{0.6876} \\ \bottomrule
\end{tabular}
}
\end{table}

We also evaluate the quality of explanations generated by LLM-based methods with the human evaluation via Amazon Mechanical Turk (AMT).
Following the convention~\cite{kim-etal-2024-pearl}, we randomly sample 100 examples from each dataset, and ask three human judges for each example to compare the quality of explanations based on the five distinct criteria\footnote{We adopt the evaluation perspectives commonly used in conversational recommender systems in terms of verifying whether an explanation has sufficient explainability to the actual users~\cite{guo2023towards}.}:
\textbf{Persuasiveness}, which measures how convincing the explanation is in making the user accept the recommendation; \textbf{Reasonability}, assessing the logical and reasonable nature of the explanation; 
\textbf{Scrutability}, determining how well the explanation helps the user understand their preferences for the item; \textbf{Faithfulness}, evaluating how accurately the explanation reflects the actual content, features, or attributes of the recommended item; and 
\textbf{Overall}, which judges the general quality of the explanation as a recommendation explanation.


In Figure~\ref{fig:humaneval}, \proposed beats all other baselines by a large margin for all criteria.
Particularly, the explanations from \proposed show remarkable \textit{persuasiveness}, implying that step-by-step reasoning based on subjective information in user/item profiles and objective information in item description exhibits great explainability.
In addition, a qualitative comparison of the explanations generated by various LLM-based recommenders in Table~\ref{tbl:casestudy} clearly supports that our textual reasoning provides more insightful and faithful explanations about the item. 
For instance, while \pfive offers a brief and vague statement and \xrec provides an abstract explanation, \proposed provides a rich, generalized explanation.
This detailed elucidation captures a variety of aspects and gives reasonable rationale for personalized recommendations that align closely with the user’s preferences.
This allows users to get a more thorough and convincing understanding of the explanations.

\subsection{Comprehensive Analysis of \proposed (RQ4)}
\label{subsec:rq4}


\begin{table}[t]
\centering
\caption{Effect of step-by-step reasoning (``R'') and bias (``B'') in rating prediction (*: $p$-value < 0.05).}
\label{tbl:no_bias_reasoning}
\resizebox{0.99\linewidth}{!}{
\begin{tabular}{cccQQQQ}
\toprule
\multirow{2.5}{*}{\textbf{Methods}} & \multirow{2.5}{*}{\textbf{R}} & \multirow{2.5}{*}{\textbf{B}} & \multicolumn{2}{c}{\textbf{IMDB}} & \multicolumn{2}{c}{\textbf{Amazon-Book}} \\ \cmidrule(lr){4-5}\cmidrule(lr){6-7}
& & & \textbf{RMSE} & \textbf{MAE} & \textbf{RMSE} & \textbf{MAE} \\ \midrule

 & \cmark & \cmark & 2.2088 & 1.6290 & 0.9100 & 0.7139 \\
\llmrec & \cmark & & 2.4221 & 1.7449 & 0.9291 & 0.7372 \\
(\gptthree, ZS) &  & \cmark & 2.2023 & 1.6519 & 0.9193 & 0.6392 \\ 
& & & 2.4843 & 1.7559 & 0.8974 & 0.6356 \\ \midrule
& \cmark & \cmark & 2.1600 & 1.5913 & \underline{0.7815} & 0.4982 \\
\llmrec & \cmark &  & 2.2254 & 1.6403 & 0.7872 & 0.4776 \\
(\gptthree, FS) & & \cmark & 2.2243 & 1.6726 & 0.8051 & 0.5246 \\ 
 & & & 2.2161 & 1.6223 & 0.7852 & \underline{0.4562}\\ \midrule
\multirow{3.5}{*}{\shortstack{\gptthree, ZS\\ (\proposed input)}} & \cmark & \cmark & 2.0688 & 1.6489 & 0.9038 & 0.6518 \\
& \cmark &  & 2.2785 & 1.8321 & 1.2453 & 1.0354 \\
& & \cmark & \underline{2.0021} & \underline{1.5789} & 0.8888 & 0.5562 \\ \midrule
& \cmark & \cmark & \textbf{1.9122}* & \textbf{1.4691}* & \textbf{0.6508}* & \textbf{0.4297}* \\
\proposed & \cmark &  & 2.1802 & 1.6848 & 0.8793 & 0.5598 \\
& & \cmark & 2.0655 & 1.6919 & 0.9181 & 0.8006 \\ \bottomrule
\end{tabular}
}
\end{table}

\smallsection{Evaluating the effect of flexible item profile construction}
We explore various item profile construction strategies (Section~\ref{subsec:inference}) and examine how these different strategies impact the performance of \proposed. 
This flexibility allows us to customize the item profile based on specific criteria, such as helpfulness-based and neighbor-based selection. 
Table~\ref{tbl:item_synopsis_rating_prediction_table} provides detailed insights into these aspects. For example, while the helpfulness-based and neighbor-based strategies offer useful alternatives for different recommendation tasks, they are slightly less effective compared to the default setting. The default setting, which utilizes all available reviews, proves to be the most effective. 
Therefore, we set the default setting to utilize all reviews, constructing the item profile in a comprehensive manner.



\begin{figure}[h]
\centering
    \includegraphics[width=\linewidth]{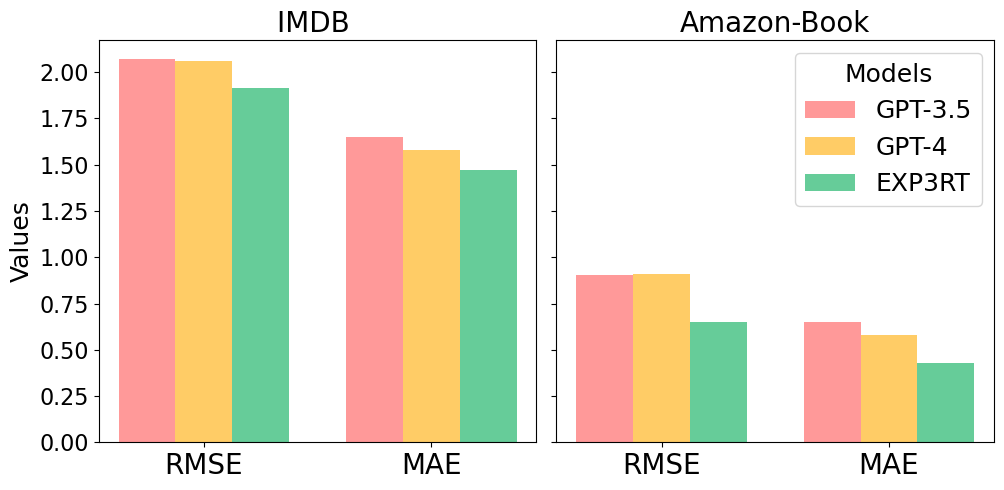}
    \caption{Validation on different LLMs with \proposed input.}
    \label{fig:diff_LLM} 
\end{figure}

\smallsection{Evaluating the effectiveness of bias and reasoning in rating prediction}
We examine the impact of user/item rating bias and textual reasoning on the performance of \proposed in rating prediction tasks by ablating these technical components. 
Additionally, to assess the impact of incorporating bias and reasoning into the approach, we evaluate the effectiveness of zero-shot and few-shot prompting-based rating prediction with \llmrec (\gptthree).
We analyze enhancements achieved by integrating (1) user/item rating bias, (2) textual reasoning, and (3) a combination of both approaches.
In Table~\ref{tbl:no_bias_reasoning}, it is obvious that both rating bias and textual reasoning are significantly helpful for reducing prediction errors.
In addition, we provide same inputs as \proposed to \gptthree as zero-shot, to show the effectiveness of knowledge distillation.
Notably, our model, which is specifically fine-tuned on the target dataset through distillation, outperforms the prompting-based approach, highlighting the necessity of LLM fine-tuning for recommendation tasks.

\smallsection{Validating performance across different LLMs}
Moreover, to validate the performance using different LLMs, we use the same prompts provided to \proposed, but employing GPT-4 instead of GPT-3.5.
The result in Figure \ref{fig:diff_LLM} supports the necessity of fine-tuning in the recommendation domain and shows that the knowledge distillation process improves a smaller student LLM, LLaMA3-8B, enhancing both cost-efficiency and recommendation accuracy.


\smallsection{Evaluating the computational efficiency}
We compare two settings to evaluate the computational efficiency of our multi-stage ranking pipeline: \textbf{\proposed-only}, where \proposed directly ranks all items without candidate selection, and \textbf{CF+\proposed}, where LightGCN first efficiently selects 20 candidate items, which are then re-ranked by \proposed. 
These experiments are conducted on the IMDB dataset using a single RTX 3090 GPU.
Integrating collaborative filtering for candidate selection reduces computation time by over 98\%.\footnote{The \proposed-only setting takes over 422 seconds per user, while the CF+\proposed setting completes the task in just over 7 seconds per user. }
The significant time reduction underscores the scalability and practicality of our approach in real-world scenarios.


%



\section{Conclusion}
\label{sec:conclusion}

This paper proposes \proposed, a novel LLM-based recommender for reasoning-enhanced rating prediction.
\proposed is optimized via distillation of a powerful teacher LLM for various tasks, including preference extraction, profile construction, and textual reasoning for rating prediction.
Our empirical evaluation on two datasets demonstrates that \proposed is effective in rating prediction and in top-k recommendation within a multi-stage ranking pipeline, where it serves as a reranker.
Furthermore, its textual reasoning can serve as faithful and reasonable explanations for recommendation, comprehensively capturing the subjective preference of users and items.
This study is the first work to propose a direction for fully leveraging the reasoning capabilities of LLMs for recommendation, based on rich information and detailed reasoning processes.


\begin{acks}
This work was supported by the IITP grants funded by the Korea government (MSIT) (No. RS-2020-II201361; RS-2024-00457882, AI Research Hub Project; IITP-2025-RS-2020-II201819), and the KBSI (National research Facilities and Equipment Center) grant funded by the Korea government (MSIT) (No. RS-2024-00403860).
\end{acks}

\bibliographystyle{ACM-Reference-Format}
\bibliography{base}

\appendix

\section{Prompts}
\label{sec:prompt}




\begin{table*}[htbp]
    \centering
    \caption{The prompt for extracting preference descriptions from raw reviews.}
    \label{tbl:feature_prompt}
    \begin{tabular}{P}
    \toprule
    \textbf{Prompt: Preference Extraction} \\
    \midrule
    Given a review written by a user, list about the ``preference'' the user liked and disliked about the item, under [\textit{Like}] and [\textit{Dislike}] in bullet points, respectively. 
    If there is nothing to mention about like/dislike, simply write ``None.'' under the corresponding tag. DO NOT write any content that is not revealed in the review.\\ \\
    \#\#\#Output Format:\\
    {[\textit{Like}]}\\
    - Encapsulate the ``preference'' user liked about the item in bullet points.\\
    {[\textit{Dislike}]}\\ 
    - Encapsulate the ``preference'' user disliked about the item in bullet points.\\
    \\
    Here is the review written by the user: \{review\} \\
    \bottomrule
    \end{tabular}
\end{table*}
\begin{table*}
    \centering
    \caption{The prompt for constructing a user profile based on the user's preference descriptions.}
    \label{tab:user_prompt}
    \begin{tabular}{P}
    \toprule
    \textbf{Prompt: User Profile Construction} \\
    \midrule
    These are the user's preferences about items: \{preferences\}\\ \\

    Based on this preferences, point out the personality of the user under [\textit{Like}] and [\textit{Dislike}] in bullet point, respectively. 
    If there is nothing to mention about like/dislike, simply write ``None.'' under the corresponding tag. 
    List only about the preferences. DO NOT mention ``Based on $\sim$''. \\ \\

    \#\#\# Output Format:\\
    {[\textit{Like}]}\\
    - Summarize the ``preferences'' of the user in bullet points.\\ 
    {[\textit{Dislike}]}\\
    - Summarize the ``preferences'' of the user in bullet points. \\ \\

    Each personalities should be about 5. \\
   
    \bottomrule
    \end{tabular}
\end{table*}
\begin{table*}
    \centering
    \caption{The prompt for constructing an item profile based on the item's preference descriptions.}
    \label{tab:item_prompt}
    \begin{tabular}{P}
    \toprule
    \textbf{Prompt: Item Profile Construction} \\
    \midrule
    These are users' preferences about the item : \{preferences\} \\ \\

    Based on this preferences, point out the ``preference'' people liked and disliked about the item under [\textit{Like}] and [\textit{Dislike}] in bullet point, respectively. 
    If there is nothing to mention about like/dislike, simply write ``None.'' under the corresponding tag. 
    List only about the preferences. DO NOT mention ``People'' or ``They''.\\ \\

    \#\#\# Output Format:\\
    {[\textit{Like}]}\\
    - Summarize the ``preference'' people liked about the item in bullet points.\\
    
    [\textit{Dislike}]\\
    - Summarize the ``preference'' people disliked about the item in bullet points.\\ \\ 

    Make sure each preferences are in a completed form. Generate less then 150 tokens.\\
   
    \bottomrule
    \end{tabular}
\end{table*}
\begin{table*}
    \centering
    \caption{The prompt for generating step-by-step reasoning on rating prediction.}
    \label{tab:reasoning_prompt}
    \begin{tabular}{P}
    \toprule
    \textbf{Prompt: Step-by-step Reasoning Generation} \\
    \midrule
    You are a recommender who provides a reason for whether or not to recommend a specific item to the user.
    You will be given a <User Rating>, <User Profile>, <Item Description>, and an <Item Profile>. 
    Based on this features, create a explanation of whether or not to recommend it. Only make a recommendation in case of <User Rating> of ``3'' or higher. \\ \\

    Let's think step by step. \\
    1. Think about the user's preference based on <User Profile>. \\
    2. Connect the user profile with item description and item profile, in which specific point the user will like/dislike this item. \\
    3. Create explanation of recommendation based on the given information.\\ \\

    Follow the instructions : \\
    - Rationale should contain only about ``the reason'' why user would like/dislike this item. \\
    - Assume that the user have never seen the item. \\
    - <User Profile> SHOULD NOT BE directly revealed in the explanation. \\
    - You must not mention that you have refer to the given information. \\
    - Do not just list the preferences, and make sure your explanation have causality. \\
    - Rationale should be able to convince users. \\ 
    - It should include a description of how it will suit the user's taste or how it will not suit the user's taste. \\
    - Generate about 5 sentences. \\
    - Mention the user as ``the user''. DO NOT mention you as ``I'' \\
    - You are recommending to one user. DO NOT mention the user as ``they''.\\
    - DO NOT start with ``Based on ~~''.\\
    ** DO NOT start with ``I would ~~''. \\ \\

    Here is the rating of item by the user:\\
    <User Rating>\\
    \{user rating\}\\ \\
    
    Here are some information about user's preferences:\\
    <User Profile>\\
    \{user profile\}\\ \\
    
    Some features of the item you should know:\\
    <Item Description>\\
    \{item description\}\\ \\
    
    <Item Profile>\\
    \{item profile\}\\ 
   
    \bottomrule
    \end{tabular}
\end{table*}
\begin{table*}
    \centering
    \caption{The prompt for Personalized Preference Reasoning-based Rating Prediction in \imdb dataset. For \amazon dataset, we use same prompt with different rating scale.}
    \label{tab:reasoning_expert}
    \begin{tabular}{P}
    \toprule
    \textbf{Prompt: Training Prompt for Rating Prediction.} \\
    \midrule
    \textbf{INPUT:}\\
    <|SYSTEM|>\\
    You are a helpful AI assistant for item recommendation. \\
    Based on the user's preferences and item characteristics provided, generate a recommendation reasoning and predict the user's rating.\\
    You must always generate a response in the following format whenever the user provides information: \\ \\
    
    Reasoning: [Provide a detailed, single-paragraph reasoning for your prediction, addressing at least three specific points of alignment or misalignment between the user's preferences and the item's characteristics.]\\
    Predicted User Rating: [Predict the user's rating as an integer from 0 to 9: 0, 1, 2, 3, 4, 5, 6, 7, 8, 9. 0 indicates the user would strongly dislike the item, while 9 indicates the user would highly enjoy and recommend it. Consider the average ratings provided for the user and the item in your prediction.]\\
    Note: Do not simply repeat the input text. Generate a new reasoning and rating prediction based on the input provided.\\ \\

    <|USER|>\\            
    I need a recommendation for this item. Here's the information: \\ \\
    User Preferences:\\
    <User Profile>\\
    \{user profile\}\\ \\
    Item Characteristics:\\
    <Item Description>\\
    \{item description\}\\ \\
    <Item Profile>\\
    \{item profile\}\\
    Based on this information, please provide a detailed reasoning for your recommendation and predict a rating for this item.\\
    Follow the format specified in the system instructions. \\ \\ 

    \textbf{TARGET:}\\
    <|ASSISTANT|>\\     
    Reasoning: \{reasoning\} \\ 
    Predicted User Rating: \{rating\}\\
    \bottomrule
    \end{tabular}
\end{table*}

The prompts play an integral role in facilitating the model's operation across the three primary steps. These prompts are designed to systematically guide each phase of the process.

\smallsection{Step 1: Preference Extraction and Encapsulation}  
The prompt is crafted to direct the model in identifying and extracting key preferences from raw reviews, subsequently organizing these into structured and interpretable sets, ensuring clarity and consistency.
The prompts can be found in Table~\ref{tbl:feature_prompt}.

\smallsection{Step 2: User and Item Profile Construction}  
This prompt is employed to aggregate and synthesize the extracted preferences into comprehensive user and item profiles, effectively summarizing relevant information for subsequent computational tasks. 
The prompts for user profiles can be found in Table~\ref{tab:user_prompt} and the prompts for item profiles can be found in Table~\ref{tab:item_prompt}.

\smallsection{Step 3: Rating Prediction through Textual Reasoning}  
The final prompt is formulated to guide the model in constructing a reasoning process that predicts user ratings. This involves integrating the generated profiles with supplementary item descriptions to derive accurate and interpretable predictions.
The prompts for generating reasoning with the teacher LLM can be found in Table~\ref{tab:reasoning_prompt}, and the prompts used for our model’s training and inference can be found in Table~\ref{tab:reasoning_expert}.




\section{User Similarity Calculation}
\label{sec:item_synopsis_construction}

To effectively construct item profiles in scenarios requiring neighbor-based selection, we calculate user-user cosine similarity using the following process. 
This approach identifies users with similar preferences, enabling personalized and context-aware recommendations.
(1) \textbf{Data Loading and Preprocessing:} User-item rating data is loaded from training data and organized into a user-item rating matrix \( R \), where \( R_{ui} \) denotes the rating of user \( u \) for item \( i \). Missing ratings are set to 0. 
(2) \textbf{Mean-Centering Ratings:} Ratings are normalized by subtracting each user's mean rating \( \mu_u \): \( R'_{ui} = R_{ui} - \mu_u \). 
(3) \textbf{Adjusted Cosine Similarity:} The similarity between users \( u \) and \( v \) is computed as 
\( \text{sim}(u, v) = \frac{\sum_{i \in I_{uv}} (R'_{ui} \cdot R'_{vi})}{\sqrt{\sum_{i \in I_{uv}} (R'_{ui})^2} \cdot \sqrt{\sum_{i \in I_{uv}} (R'_{vi})^2}} \), 
where \( I_{uv} \) is the set of items rated by both users. 
(4) \textbf{Similarity Matrix Construction:} Similarities are stored in matrix \( S \), where \( S_{uv} \) represents similarity between users \( u \) and \( v \).
(5) \textbf{Identifying Top Similar Users:} For each test interaction, the top 3 most similar users from the training set who interact with the same item are selected based on \( S \). 
(6) \textbf{Preference Set and Item Profile:} Reviews from the top 3 similar users are aggregated to form a preference set, which is used to construct the item profile.

\section{Implementation Details}
\label{sec:implementation_detail}
\smallsection{Baselines}
We use publicly available code for the baselines and strictly follow the preprocessing and implementation steps instructed for the code. 
In cases where official codes are not provided (e.g., \mf~\cite{Koren2009MatrixFT}, \mlp~\cite{Cheng2016WideD}), we prepare the rating matrices for our datasets, \imdb and \amazon, 
then implement methods as described in the corresponding papers.
Especially for \pfive~\cite{Geng2022RecommendationAL}, we fine-tune T5 on each dataset using prompts aligned with our tasks.
For rating prediction task, we use all prompts whose target template is \texttt{star\_rating}, then report the average score. 
For explanation generation task, we utilize the prompt with ID 3-1.






\end{document}